\documentclass[11pt]{article}
\usepackage{float}
\usepackage[table]{xcolor}
\usepackage{multirow}
\usepackage{xfp}
\usepackage{amssymb}
\usepackage{makecell}
\usepackage[preprint]{acl}
\usepackage{booktabs}
\usepackage{times}
\usepackage{latexsym}
\usepackage{comment}
\usepackage{algpseudocode}
\usepackage[section]{placeins}
\usepackage{amsmath}
\usepackage[T1]{fontenc}
\usepackage[utf8]{inputenc}

\usepackage{microtype}

\usepackage{inconsolata}

\usepackage{graphicx}
\usepackage[linesnumbered,ruled,vlined]{algorithm2e}

\newcommand{\algcmt}[1]{\tcp{\textcolor{gray}{#1}}}

\newcommand{\ours}{SkillBrew}

\title{\ours: Multi-Objective Curation of Skill Banks for LLM Agents}

\author{
Wentao Hu\textsuperscript{\rm $\spadesuit$}\thanks{Equal contribution.},
Zhendong Chu\textsuperscript{\rm $\heartsuit$}\thanks{Corresponding author.}\footnotemark[1],
\textbf{
Yiming Zhang\textsuperscript{\rm $\clubsuit$},
Junda Wu\textsuperscript{\rm $\diamondsuit$},
Ming Jin\textsuperscript{\rm $\sigma$}},\\
\textbf{
Xiangyu Zhao\textsuperscript{\rm $\spadesuit$},
Yilei Shao\textsuperscript{\rm $\triangle$},
Yanfeng Wang\textsuperscript{\rm $\square$},
Qingsong Wen\textsuperscript{\rm $\heartsuit$}}\\
\textsuperscript{\rm $\spadesuit$}City University of Hong Kong
\textsuperscript{\rm $\heartsuit$}Squirrel Ai Learning \\
\textsuperscript{\rm $\clubsuit$}University of Science and Technology of China
\textsuperscript{\rm $\diamondsuit$}University of California, San Diego \\
\textsuperscript{\rm $\sigma$}Griffith University
\textsuperscript{\rm $\triangle$}East China Normal University
\textsuperscript{\rm $\square$}Shanghai Jiao Tong University \\
\texttt{\{zc9uy@virginia.edu\}}
}

\begin{document}
\maketitle

\begin{abstract}
Retrieval-augmented LLM agents increasingly rely on curated skill banks: collections of reusable textual principles that guide decision making on complex tasks. Existing approaches typically expand these banks in an append-only fashion, continuously adding new skills without removing redundant, outdated, or harmful ones, resulting in inefficient and poorly curated repositories. In this paper, we formulate skill bank curation as a constrained multi-objective problem: a desirable bank must be \emph{useful} for the agent, \emph{diverse} in its content, and provide good \emph{coverage} of the query distribution. To this end, we introduce \textbf{\ours}, a multi-objective curation framework that formalizes skill bank curation as Pareto-aware optimization under a utility constraint, and solves it via a bi-level propose-then-verify loop. We evaluate our approach on two public benchmarks. Our findings suggest that treating skill banks as objects of principled curation, rather than ever-growing append-only logs, is an important step toward building self-improving LLM agents.

\end{abstract}

\section{Introduction}

Large language model (LLM) agents \citep{yao2022react,shinn2023reflexion} are increasingly deployed on long-horizon, open-ended tasks where the ability to accumulate and reuse procedural knowledge across episodes determines downstream performance. A widely adopted abstraction is the \emph{skill bank}: a retrievable collection of distilled, reusable textual or executable artifacts that encode how the agent performs recurring tasks \citep{wang2023voyager,zhao2024expel,ni2026trace2skill}. This abstraction has been standardized by Anthropic's Agent Skills specification \citep{anthropic2025skills} and is now adopted by modern agent harnesses such as Claude Code. However, the skill bank is not a one-shot artifact: a bank that is simultaneously useful, diverse, and well-covered cannot be obtained by a single distillation pass, and requires iterative refinement that balances these competing properties \citep{wang2023voyager,fang2025memp}. Yet how to build, refine, and prune skills jointly as a bank, rather than judging each skill in isolation, remains largely unexamined.

\begin{figure}
    \centering
    \includegraphics[width=1\linewidth]{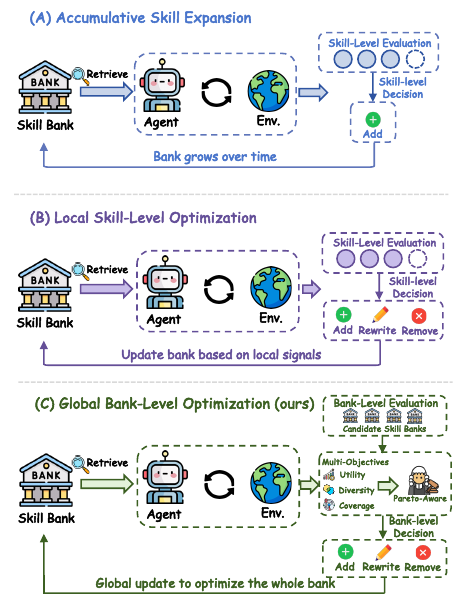}
    \caption{Three paradigms of skill bank curation. Prior work either (A) accumulates skills without removal, or (B) edits skills based on per-skill signals. We propose (C) global bank-level optimization, which evaluates the entire bank under multi-objective trade-offs and chooses updates via a Pareto-aware selector.}
    \label{fig:motivation}
\end{figure}

A growing line of work tackles skill bank construction, which we organize into two paradigms (Figure~\ref{fig:motivation}). The first, \emph{accumulative skill expansion}, continuously adds new skills from experience and lets the bank grow over time, as in Voyager's self-verified skill library \citep{wang2023voyager}, Trace2Skill's hierarchical consolidation of trajectory-local lessons \citep{ni2026trace2skill}, and SkillRL's failure-driven skill evolution coupled with RL policy updates \citep{xia2026skillrl}. The second, \emph{local skill-level optimization}, retains, edits, or removes skills based on a single scalar curation signal, such as ExpeL's vote-based \textsc{Add}/\textsc{Edit} operations on cross-task insights~\citep{zhao2024expel}, top-$k$ frontier selection on held-out accuracy~\citep{alzubi2026evoskill}, and non-parametric scoring gates~\citep{mi2026skill}. While these methods advance skill bank construction along different axes, their curation reduces to a single scalar criterion, and bank-level properties such as diversity and coverage remain a less explored dimension.


Principled skill bank curation is challenging for three interrelated reasons. First, the quality of a bank is inherently \emph{multi-dimensional}: a good bank must be useful, internally diverse, and broadly cover the query distribution. These criteria are mutually constraining: adding skills to broaden coverage tends to introduce redundancy that lowers diversity, while pruning to preserve diversity reduces coverage; they cannot be collapsed into a single scalar without introducing pathologies. Second, \emph{credit assignment} is difficult: the contribution of any individual skill is observable only through expensive rollouts, and is further confounded by interactions with co-retrieved skills, making it hard to attribute outcomes to specific entries. Third, the objectives are \emph{non-differentiable and expensive to query}, while the space of admissible edits is open-ended, since new skills are synthesized from trajectory evidence rather than chosen from a fixed pool. This rules out gradient-based optimization and exhaustive search, so curation requires a mechanism that can both propose plausible edits and verify them under the multi-objective criterion. These challenges explain why prior work has defaulted to accumulation without pruning or single-scalar curation rather than principled multi-objective curation.


In this paper, we propose \textbf{\ours}, a training-free framework for iterative skill bank refinement that introduces a third paradigm, \emph{global bank-level optimization} (Figure~\ref{fig:motivation}C), and addresses the above challenges through three design choices. First, we formalize skill bank curation as \emph{constrained multi-objective optimization} over utility, diversity, and coverage, with utility as an explicit constraint and diversity/coverage as structural regularizers, reflecting the asymmetry that a bank must first be useful before its organization matters. Second, we attribute credit through \emph{per-skill counterfactual replay}: leave-one-out rollouts isolate the marginal contribution of each skill, providing fine-grained \textsc{Keep}/\textsc{Rewrite}/\textsc{Remove} evidence for bank-level optimization. Third, we navigate the non-differentiable and open-ended edit space via a \emph{bi-level propose-then-verify loop}, which uses trajectory evidence on a support split to propose candidate edits and Pareto-aware selection on a held-out query split to verify them, preventing overfitting to the trajectories that motivated each edit. 

Across two public benchmarks, \textbf{\ours} substantially outperforms strong baselines and transfers across worker LLMs of different scales, suggesting that treating skill banks as objects of principled curation, rather than judging skills in isolation, yields more effective procedural knowledge for retrieval-augmented agents.

\section{Related Work}

\subsection{From Agent Memory to Agent Skill}
A central challenge in deploying LLM agents~\citep{yao2022react} on long-horizon tasks is that each task is typically treated as an isolated episode, leaving no straightforward mechanism to accumulate transferable knowledge across tasks. One line of work equips the agent with a persistent memory of past interaction traces~\citep{chhikara2025mem0,fang2025memp}, retrieved at inference time as references for similar future tasks. 

A distinct line abstracts raw experience into skills: compact and reusable procedural artifacts that capture how agents perform recurring tasks. Voyager~\citep{wang2023voyager} first brought this paradigm to LLM agents by maintaining a growing library of executable code skills in Minecraft, and subsequent work generalizes the formulation to textual forms, including cross-task insights distilled from trajectories~\citep{zhao2024expel} and transferable skills consolidated from trajectory pools via parallel hierarchical analysis~\citep{ni2026trace2skill}. This paradigm has further been standardized by Anthropic's Agent Skills specification~\citep{anthropic2025skills}. Across these designs, the skill bank serves as a compact retrievable representation of procedural knowledge, but how to curate this bank iteratively, balancing the utility of individual skills against the diversity and coverage of the bank as a whole, remains largely open.

\subsection{Skill Bank Curation}
Early work on skill libraries, exemplified by Voyager~\citep{wang2023voyager}, treats the bank as an append-only artifact that grows as the agent succeeds, without mechanisms to refine or prune stored skills. More recent work introduces explicit curation over individual skills. ExpeL~\citep{zhao2024expel} updates cross-task insights through vote-based ADD/EDIT operations, CoEvoSkills~\citep{zhang2026coevoskills} uses a surrogate verifier for failure feedback, Skill-Pro~\citep{mi2026skill} admits and prunes skills through online scoring, and SkillForge~\citep{liu2026skillforge} closes a failure-driven refinement loop. However, the curation criterion in these methods collapses to a single scalar signal, folding utility, diversity, and coverage of the bank into one ordering, so the trade-offs that arise during curation cannot in principle be resolved.

\section{Methodology}
\label{sec:method}
\begin{figure*}[t]
    \centering
    \includegraphics[width=\textwidth]{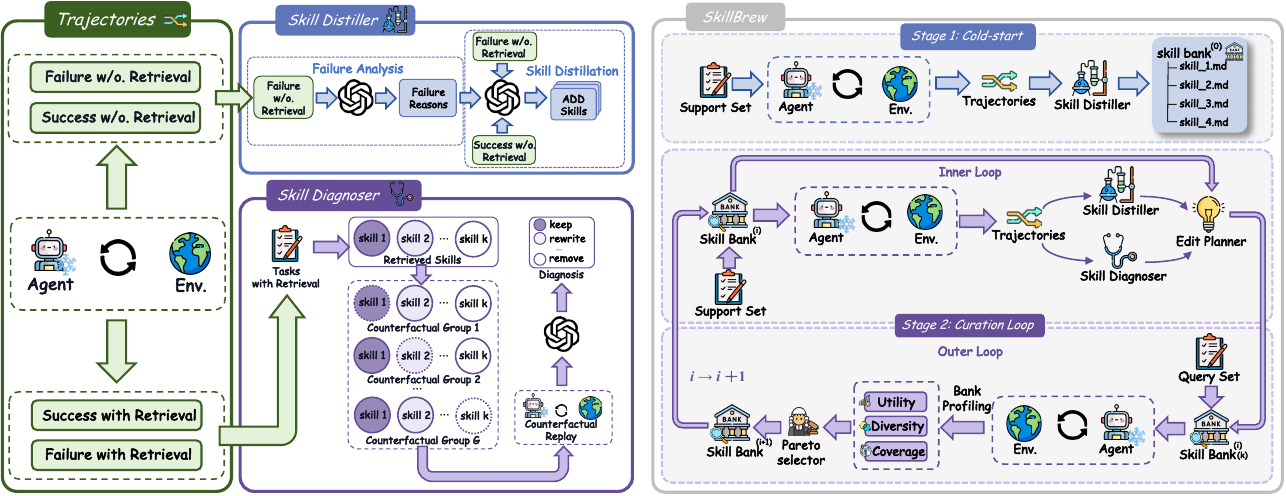}
    \caption{Overview of {\ours}. At each round, an inner loop produces \textsc{Add} candidates via the \emph{Skill Distiller}, \textsc{Keep}, \textsc{Rewrite}, and \textsc{Remove} verdicts via the \emph{Skill Diagnoser}, and the \emph{Edit Planner} sees current skill bank and proposes $K$ new candidate banks from these edits. An outer loop then verifies these candidate banks on the query set under three objectives $J_{\text{util}}$, $J_{\text{div}}$, and $J_{\text{cov}}$, and selects the next bank with a Pareto-aware selector.}
    \label{fig:overview}
\end{figure*}

In this section, we present \textbf{\ours}, a training-free framework for principled skill bank curation in LLM agents, as illustrated in Figure~\ref{fig:overview}. We first introduce the core principles underlying effective skill bank construction, formalizing the desiderata of a curated bank in terms of \emph{utility}, \emph{diversity}, and \emph{coverage}. We then cast skill bank curation as a constrained multi-objective optimization problem, where utility serves as the primary constraint while diversity and coverage regularize the bank structure. To solve this problem, we develop a bi-level propose-then-verify loop that iteratively generates candidate edits from trajectory evidence and selects non-dominated banks under a utility constraint.

\subsection{Problem Definition}
\label{sec:problem_definition}
 Let $\mathcal{S}$ denote the space of admissible skills, where each skill $s \in \mathcal{S}$ is a piece of reusable procedural knowledge that can be retrieved to guide agent behavior. A skill bank is defined as a subset $\mathcal{B} \subseteq \mathcal{S}$ serving as the retrieval pool for a frozen agent $\pi_{\text{agent}}$.

Given a task query $t$, a retriever $\textsc{ret}$ returns a subset of relevant skills $R_t = \textsc{ret}(\mathcal{B}, t) \subseteq \mathcal{B}$, and the agent rolls out a trajectory
\begin{equation}
\tau_t = \pi_{\text{agent}}(t \mid R_t)
\end{equation}
conditioned on $R_t$.

A desirable skill bank should satisfy three complementary principles. First, retrieved skills should be \emph{useful}, meaning they improve the agent's task performance. Second, the bank should be \emph{diverse}, avoiding redundant or overly similar skills. Third, the bank should provide broad \emph{coverage} over the query distribution so that useful skills can be retrieved across a wide range of tasks. These principles are in tension: a bank of mutually orthogonal skills attains high diversity but offers little utility if the skills are off-task; aggressively pruning to a few high-utility skills sacrifices coverage on the long tail. We therefore cast skill bank curation as a \textbf{constrained multi-objective optimization}:
\begin{equation}
\label{eq:cmoo}
\begin{gathered}
\mathcal{B}^{\star}
\in
\operatorname*{Pareto\text{-}arg\,max}_{\mathcal{B} \in \mathcal{F}}
\Big(
J_{\mathrm{util}}(\mathcal{B}),
J_{\mathrm{div}}(\mathcal{B}),
J_{\mathrm{cov}}(\mathcal{B})
\Big), \\
\mathcal{F}
=
\left\{
\mathcal{B} \subseteq \mathcal{S}
\,\middle|\,
J_{\mathrm{util}}(\mathcal{B}) \ge \eta
\right\}.
\end{gathered}
\end{equation}
where $J_{\mathrm{util}}$, $J_{\mathrm{div}}$, and $J_{\mathrm{cov}}$ respectively measure utility, diversity, and coverage, and $\eta$ is a minimum utility threshold. The solution $\mathcal{B}^{\star}$ is Pareto-optimal~\citep{censor1977pareto}: no feasible bank improves one objective without sacrificing another.


Importantly, we treat utility as the primary constraint, while diversity and coverage act as structural regularizers over the bank. This reflects the asymmetry of the three principles: a skill bank must first be useful to the agent, while diversity and coverage improve its organization, robustness, and generalization. In the next section, we describe the computation of the three objectives in detail.

We partition the task distribution into a support set $\mathcal{D}_{\text{support}}$ and a query set $\mathcal{D}_{\text{query}}$. In this paper, all three objectives $\Big(J_{\text{util}}, J_{\text{div}}, J_{\text{cov}}\Big)$ are computed from agent rollouts on $\mathcal{D}_{\text{query}}$, while $\mathcal{D}_{\text{support}}$ is reserved for proposing candidate edits in Section~\ref{sec:optimization}. This separation prevents the bank from overfitting to the trajectories that motivated its edits.

\subsection{Principles of Skill Bank Curation}
\label{sec:principles}
We consider three principles that capture complementary and potentially conflicting aspects of skill bank quality: \emph{utility} measures whether retrieved skills improve agent performance, \emph{diversity} measures redundancy within the bank, and \emph{coverage} measures how broadly the bank's skills are retrieved and exercised across tasks.
\paragraph{Utility.}

We measure utility through per-skill counterfactual contribution. For a trajectory $\tau \in \mathcal{D}_{\text{query}}$ retrieving skill $s$, we replay the agent with $s$ removed under the same environment, yielding
\begin{equation}
\Delta(\tau, s) = r(\tau, s) - r(\tau, \emptyset),
\end{equation}
where $r(\tau, \cdot) \in [0, 1]$ is the task-level reward (success indicator or normalized score) obtained under the specified retrieval. The utility of $s$ is computed by averaging over all trajectories retrieving it, 
\begin{equation}
J_{\text{util}}(s)
=
\frac{1}{|\mathcal{T}_s|}
\sum_{\tau \in \mathcal{T}_s}
\Delta(\tau, s),
\end{equation}
where $\mathcal{T}_s = \{\tau : s \in \mathrm{ret}(\mathcal{B}, \tau)\}$. We then aggregate skill utilities into a bank-level objective, 
\begin{equation}
J_{\text{util}}(\mathcal{B})
=
\sum_{s \in \mathcal{B}} w_s \cdot J_{\text{util}}(s),
\end{equation}
where $w_s = |\mathcal{T}_s| / N_R$ and $N_R$ is the number of trajectories with non-empty retrieval.

\paragraph{Diversity.}
We measure diversity using the geometric volume spanned by skill embeddings. Let $G$ denote the Gram matrix of L2-normalized skill embeddings in $\mathcal{B}$. This objective is defined as, 
\begin{equation}
J_{\text{div}}(\mathcal{B})
=
\det(G + \varepsilon I)^{1/|\mathcal{B}|}
\in (0,1],
\end{equation}
where $\varepsilon I$ is a small diagonal regularizer for numerical stability. This formulation penalizes redundant or highly correlated skills, assigning higher scores to banks with structurally diverse skills.

\paragraph{Coverage.}
We measure coverage as the product of retrieval density and skill usage:
\begin{equation}
J_{\text{cov}}(\mathcal{B})
=
\mathrm{density}
\cdot
\mathrm{usage} \in [0, 1].
\end{equation}
The density term measures the average fraction of retrieval slots filled across tasks in $\mathcal{D}_{\text{query}}$:
\begin{equation}
\mathrm{density}
=
\frac{1}{|\mathcal{D}_{\text{query}}|}
\sum_{t \in \mathcal{D}_{\text{query}}}
\frac{|\mathrm{ret}(\mathcal{B}, t)|}{k_{\text{top}}}.
\end{equation}

The usage term measures the fraction of skills retrieved at least once:
\begin{equation}
\mathrm{usage}
=
\frac{
|\{s \in \mathcal{B} : s \text{ retrieved at least once}\}|
}{
|\mathcal{B}|
}.
\end{equation}

Together, the two terms encourage banks that not only retrieve skills frequently across tasks, but also utilize a broad portion of the bank rather than relying on a small subset of dominant skills.

\subsection{Pareto-Aware Skill Bank Curation}
\label{sec:optimization}

Directly solving Eq.~\ref{eq:cmoo} is intractable for two reasons: $\mathcal{S}$ itself is open-ended, with new skills synthesized from trajectory evidence rather than drawn from a fixed pool; and all objectives are only observable through expensive agent rollouts. We therefore propose a bi-level propose-then-verify framework that iteratively refines the skill bank through candidate generation and Pareto-aware selection.

\paragraph{Bi-level propose-then-verify loop.}
As shown in Figure~\ref{fig:overview}, at iteration $i$, given the current bank $\mathcal{B}^{(i)}$, the framework alternates between an inner loop that proposes candidate edits and an outer loop that verifies candidate banks:
\begin{equation}
\label{eq:bilevel}
\begin{gathered}
\underbrace{
\mathcal{K}^{(i)}
=
\mathrm{Plan}\!\big(
    \texttt{Distill},
    \texttt{Diagnose};
    \mathcal{B}^{(i)},
    \mathcal{D}_{\text{support}}
\big)
}_{\textbf{inner loop: candidate generation}}
\\[4pt]
\underbrace{
\mathcal{B}^{(i+1)}
=
\mathrm{Select}\!\big(
    \mathcal{K}^{(i)};
    \eta,
    \mathcal{D}_{\text{query}}
\big)
}_{\textbf{outer loop: Pareto-aware selection}}.
\end{gathered}
\end{equation}

The support/query split introduced in Section~\ref{sec:problem_definition} aligns naturally with this bi-level structure: $\mathcal{D}_{\text{support}}$ supplies trajectory evidence to the inner loop, while $\mathcal{D}_{\text{query}}$ is reserved for the outer loop's Pareto-aware selection.

\paragraph{Cold-start initialization.}
As illustrated in the cold-start stage of Figure~\ref{fig:overview}, we first construct an initial bank $\mathcal{B}^{(0)}$ from trajectories collected without retrieval. Following~\citep{zhao2024expel}, successful and failed trajectories are passed to a \emph{Skill Distiller}, which synthesizes an initial set of \textsc{Add}-type skills from failure patterns in conjunction with successful trajectories.

\paragraph{Inner loop: candidate generation.}
As illustrated in the inner loop of Figure~\ref{fig:overview} and summarized in Algorithm~\ref{alg:inner_loop}, at each iteration the inner loop transforms $\mathcal{B}^{(i)}$ into $K$ candidate banks through three skill curation operations: \textsc{Add}, \textsc{Rewrite}, and \textsc{Remove}.

Rolling out $\pi_{\text{agent}}$ on $\mathcal{D}_{\text{support}}$ with retrieval from $\mathcal{B}^{(i)}$ yields four trajectory quadrants defined by outcome (success/failure) and retrieval (empty/non-empty):
\begin{equation}
\label{eq:quadrants}
\begin{aligned}
\mathcal{T}^{+}_{\bar{R}} &= \{\tau_t : \mathrm{succ}(\tau_t),\; R_t = \emptyset\}, \\
\mathcal{T}^{-}_{\bar{R}} &= \{\tau_t : \mathrm{fail}(\tau_t),\; R_t = \emptyset\}, \\
\mathcal{T}^{+}_{R}        &= \{\tau_t : \mathrm{succ}(\tau_t),\; R_t \neq \emptyset\}, \\
\mathcal{T}^{-}_{R}        &= \{\tau_t : \mathrm{fail}(\tau_t),\; R_t \neq \emptyset\}.
\end{aligned}
\end{equation}

The no-retrieval quadrants $\mathcal{T}^{\pm}_{\bar{R}}$ expose missing capabilities in the bank and are consumed by the \emph{Skill Distiller}, which synthesizes candidate \textsc{Add} skills by clustering failure patterns in $\mathcal{T}^{-}_{\bar{R}}$ and using $\mathcal{T}^{+}_{\bar{R}}$ as positive references.

The retrieval quadrants $\mathcal{T}^{\pm}_{R}$ provide counterfactual evidence on existing skills and are consumed by a \emph{Skill Diagnoser}. For each retrieved skill $s \in R_t$, we perform a leave-one-out replay
\begin{equation}
\tilde{\tau}_{t \setminus s}
=
\pi_{\text{agent}}
\!\left(
t \mid R_t \setminus\{s\}
\right),
\end{equation}
which isolates the marginal contribution of $s$. Based on the factual outcome of $\tau_t$ and the counterfactual outcome of $\tilde{\tau}_{t \setminus s}$ aggregated over all trajectories retrieving $s$, the \emph{Skill Diagnoser} assigns one of three verdicts: \textsc{Keep}, \textsc{Rewrite}, or \textsc{Remove}.

Finally, an \emph{Edit Planner} composes the three edit pools $(\mathcal{E}_{\text{add}}, \mathcal{E}_{\text{rewrite}}, \mathcal{E}_{\text{remove}})$ into $K$ candidate banks $\mathcal{K}^{(i)} = \{\mathcal{B}^{(i+1)}_k\}_{k=1}^{K}$, while preserving the \textsc{Keep} skills as protected slots. 

The full prompts used to implement \emph{Skill Distiller}, \emph{Skill Diagnoser}, and \emph{Edit Planner} are listed in Appendix~\ref{sec:appc}.

\paragraph{Outer loop: Pareto-aware selection.}
Given the candidate set $\mathcal{K}^{(i)}$, the outer loop evaluates each candidate by its objective profile
\begin{equation}
\Phi(\mathcal{B})
=
\big(
J_{\text{util}}(\mathcal{B}),
J_{\text{div}}(\mathcal{B}),
J_{\text{cov}}(\mathcal{B})
\big).
\end{equation}
We also include a \emph{null candidate}, which represents keeping the current bank unchanged.
This allows the selector to make no edit when none of the proposed candidates improves the bank. Over $\mathcal{K}^{(i)} \cup \{\text{null}\}$, we first compute the non-dominated Pareto front~\citep{deb2002fast} under the three objectives, then resolve the front lexicographically with $J_{\text{util}}$ as the highest priority.

Since the absolute utility threshold $\eta$ in Eq.~\ref{eq:cmoo} cannot be specified a priori over an open-ended skill space, we replace it with a \emph{front-adaptive} threshold $u_{\max}-\epsilon$.
Let $u_{\max}$ denote the largest $J_{\text{util}}$ on the front, and let
\begin{equation}
\mathrm{Tied} = \{\mathcal{B} \in \mathrm{front} : J_{\text{util}}(\mathcal{B}) \ge u_{\max} - \epsilon\}
\end{equation}
denote the candidates within tolerance $\epsilon$ of the lex-optimal utility. If $|\mathrm{Tied}| = 1$, the unique tied candidate wins; otherwise we break the tie inside $\mathrm{Tied}$ by hypervolume contribution over the two regularizers $J_{\text{div}}$ and $J_{\text{cov}}$. This yields a per-round non-degradation guarantee, $J_{\text{util}}(\mathcal{B}^{(i+1)}) \ge J_{\text{util}}(\mathcal{B}^{(i)}) - \epsilon$, by construction.

If the null candidate wins the tie-break, the bank carries forward unchanged into the next round; the loop terminates when the maximum number of rounds $T$ is reached. We refer the reader to Algorithm~\ref{alg:outer_loop} for the full procedure.

\section{Experiments}
\label{sec:exp}

\begin{table*}[!t]
\centering
\caption{Main results on ALFWorld (Success Rate, \%) and WebShop (Score and Success Rate). All methods use Qwen2.5-7B-Instruct as the frozen agent worker. \textbf{Bold} denotes best results per column; \textcolor{blue}{Blue} marks the ReAct baseline.}
\label{tab:main}
\small
\setlength{\tabcolsep}{4pt}
\begin{tabular}{l ccccccc cc}
\toprule
& \multicolumn{7}{c}{\textbf{ALFWorld (Success \%)}} & \multicolumn{2}{c}{\textbf{WebShop}} \\
\cmidrule(lr){2-8} \cmidrule(lr){9-10}
\textbf{Method} & Pick & Look & Clean & Heat & Cool & Pick2 & Avg. & Score & Succ. \\
\midrule
\rowcolor{gray!15}\multicolumn{10}{c}{\textit{Vanilla Baselines}} \\
Zero-Shot & 33.4 & 21.6 & 19.3 & 6.90 & 2.80 & 3.20 & 14.8 & 26.4 & 7.80 \\
ReAct~\citep{yao2022react} & \textcolor{blue}{48.5} & \textcolor{blue}{35.4} & \textcolor{blue}{34.3} & \textcolor{blue}{13.2} & \textcolor{blue}{18.2} & \textcolor{blue}{17.6} & \textcolor{blue}{31.2} & \textcolor{blue}{46.2} & \textcolor{blue}{19.5} \\
\rowcolor{gray!15}\multicolumn{10}{c}{\textit{Memory-Based Methods}} \\
Reflexion~\citep{shinn2023reflexion} & 62.0 & 41.6 & 44.9 & 30.9 & 36.3 & 23.8 & 42.7 & 58.1 & 28.8 \\
Mem0~\citep{chhikara2025mem0} & 54.0 & 55.0 & 26.9 & 36.4 & 20.8 & 7.69 & 33.6 & 23.9 & 2.00 \\
MemP~\citep{fang2025memp} & 54.3 & 38.5 & 48.1 & 56.2 & 32.0 & 16.7 & 41.4 & 25.3 & 6.40 \\
SimpleMem~\citep{liu2026simplemem} & 64.5 & 33.3 & 20.0 & 12.5 & 33.3 & 3.84 & 29.7 & 33.2 & 8.59 \\
\rowcolor{gray!15}\multicolumn{10}{c}{\textit{Skill-Based Methods}} \\
Voyager~\citep{wang2023voyager} & 62.5 & 61.1 & 51.6 & 52.2 & 28.6 & 17.6 & 47.0 & 50.0 & 26.4 \\
ExpeL~\citep{zhao2024expel} & 21.0 & 67.0 & 55.0 & 52.0 & \textbf{71.0} & 6.00 & 46.3 & 30.9 & 11.2 \\
EvoSkill~\citep{alzubi2026evoskill} & \textbf{70.8} & 27.8 & 41.9 & 39.1 & 19.0 & 11.8 & 37.3 & 42.2 & 12.6 \\
Skill-Pro~\citep{mi2026skill} & 66.7 & 66.7 & 45.2 & 43.5 & 47.6 & 23.5 & 49.3 & 38.7 & 23.2 \\
\midrule
\textbf{\ours} & \textbf{70.8} & \textbf{72.2} & \textbf{58.1} & \textbf{60.9} & 57.1 & \textbf{29.4} & \textbf{59.0} & \textbf{59.3} & \textbf{38.4} \\
\bottomrule
\end{tabular}
\end{table*}

\begin{table*}[!t]
\centering
\caption{Cross-worker generalization on ALFWorld (Success Rate, \%) and WebShop (Score and Success Rate). \textbf{\ours} is applied on five worker backbones, including open-source and closed-source models.}
\label{tab:cross_basemodel}
\small
\setlength{\tabcolsep}{4pt}
\begin{tabular}{l l ccccccc cc}
\toprule
& & \multicolumn{7}{c}{\textbf{ALFWorld (Success \%)}} & \multicolumn{2}{c}{\textbf{WebShop}} \\
\cmidrule(lr){3-9} \cmidrule(lr){10-11}
\textbf{Method} & \textbf{Worker} & Pick & Look & Clean & Heat & Cool & Pick2 & Avg. & Score & Succ. \\
\midrule
\multirow{5}{*}{ReAct}               & Qwen2.5-7B-Instruct          & 48.5  & 35.4  & 34.3  & 13.2  & 18.2  & 17.6  & 31.2  & 46.2  & 19.5  \\
                                     & Qwen3-4B-Instruct-2507       & 70.8  & 55.6  & 19.4  & 39.1  & 23.8  & 29.4  & 38.8  & 49.3 & 16.8 \\
                                     & Qwen3-30B-A3B-Instruct-2507  & 83.3  & 61.1  & 41.9  & 52.2  & 33.3  & 64.7  & 55.2  & 43.8  & 14.0  \\
\cmidrule(lr){2-11}
                                        & GPT-4o         & 81.2 & 63.1 & 23.2 & 56.1 & 20.2 & 41.0 & 46.4 & 33.3 & 27.2 \\
                                        & Gemini-2.5-Pro & 85.2 & 53.1 & 67.3 & 68.2 & 29.1 & 59.4 & 61.8 & 52.3 & 38.1 \\
\midrule
\multirow{5}{*}{\textbf{\ours}} & Qwen2.5-7B-Instruct          & 70.8     & 72.2     & 58.1     & 60.9     & 57.1     & 29.4     & 59.0     &59.3     & 38.4     \\
                                     & Qwen3-4B-Instruct-2507       & 79.2  & 61.1  & 54.8  & 56.5  & 66.7  & 41.2  & 60.4  & 51.0  & 36.6  \\
                                     & Qwen3-30B-A3B-Instruct-2507  & 83.3  & 77.8  & 80.6  & 82.6  & 61.9  & 58.8  & 75.4  & 58.4  & 56.2  \\
\cmidrule(lr){2-11}
                                        & GPT-4o                       & 91.7  & 88.9  & 87.1  & 91.3  & 81.0  & 88.2  & 88.1  & 75.0  & 65.0  \\
                                        & Gemini-2.5-Pro               & 95.8  & 94.4  & 90.3  & 87.0  & 76.2  & 94.1  & 89.6  & 82.4  & 72.0  \\
\bottomrule
\end{tabular}
\end{table*}

\begin{figure*}
    \centering
    \includegraphics[width=0.9\linewidth]{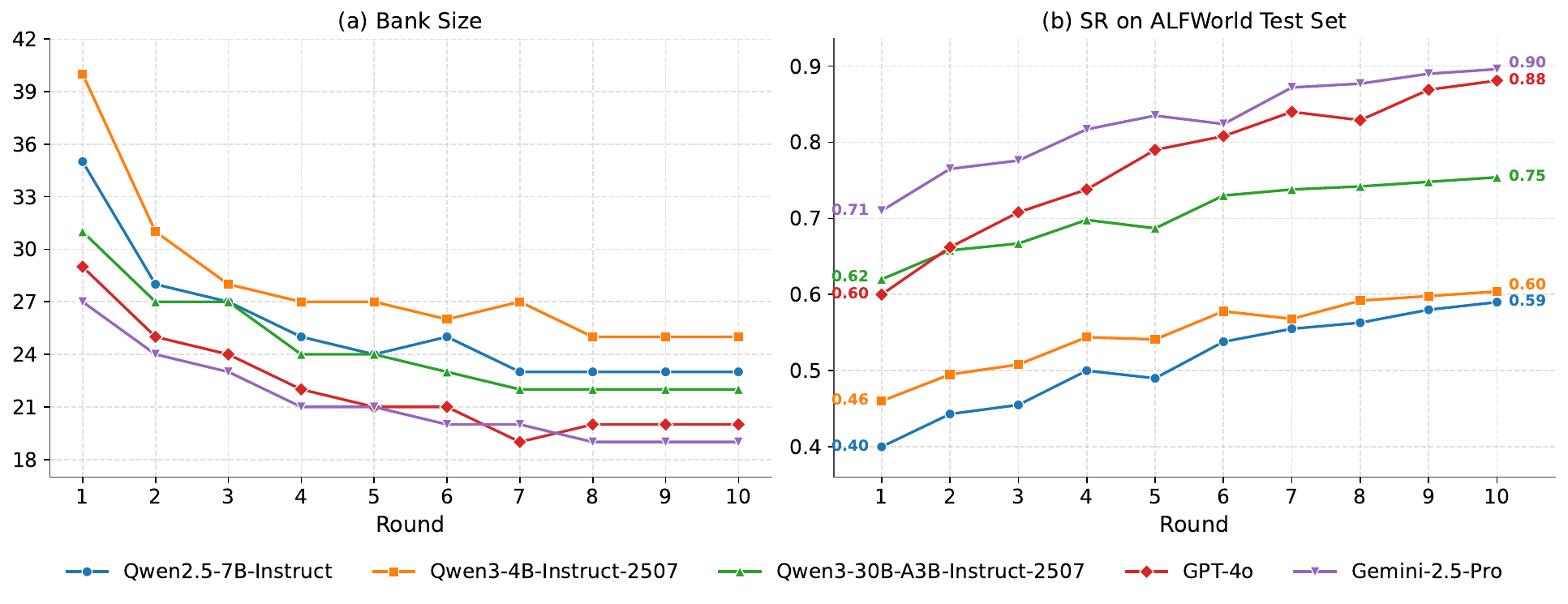}
    \caption{Curation loop dynamics of {\ours} across five worker backbones on ALFWorld. (a) Bank size across rounds. (b) Success rate on ALFWorld test set across rounds.}
    \label{fig:dynamics}
\end{figure*}

\subsection{Experimental Setup}

\paragraph{Environments.}
We evaluate on two agent benchmarks. ALFWorld~\citep{shridhar2020alfworld} is a textualized embodied benchmark spanning six household task families that require compositional action sequences; we report per-task and overall success rate (\%) on its test split. WebShop~\citep{yao2022webshop} is a simulated online shopping benchmark requiring multi-step navigation, option customization, and purchase decisions. We report two metrics on its test split: Score, the average normalized reward measuring attribute matching between the purchased and target product, and Succ., the fraction of episodes where every required attribute is matched.

\paragraph{Baselines.}
We compare {\ours} against ten baselines in Table~\ref{tab:main}, all using Qwen2.5-7B-Instruct as the frozen agent worker: (1) vanilla baselines (Zero-Shot and ReAct~\citep{yao2022react}); (2) memory-based agents (Reflexion~\citep{shinn2023reflexion}, Mem0~\citep{chhikara2025mem0}, MemP~\citep{fang2025memp}, and SimpleMem~\citep{liu2026simplemem}); and (3) skill-based methods (Voyager~\citep{wang2023voyager}, ExpeL~\citep{zhao2024expel}, EvoSkill~\citep{alzubi2026evoskill}, and Skill-Pro~\citep{mi2026skill}). These training-free baselines are evaluated under frozen-worker setting. We do not compare with training-based methods such as SkillRL, which jointly updates the worker policy and the bank, since this changes the underlying agent and is not applicable to closed-source workers. We leave joint agent-bank optimization to future work.

\paragraph{Implementation details.}
We use GPT-5.4~\citep{openai2026gpt54} for the \emph{Skill Distiller}, \emph{Skill Diagnoser}, and \emph{Edit Planner}. For ALFWorld, we sample 200 \texttt{train} episodes as $\mathcal{D}_{\text{support}}$, and use the full \texttt{valid\_seen} split as $\mathcal{D}_{\text{query}}$, and report results on the full \texttt{valid\_unseen} split as the test set. For WebShop we follow its standard test split with 500 episodes. Each skill in our bank is structured with: a concise title, a principle describing the strategy, and when\_to\_apply conditions specifying applicability. The agent worker adopts ReAct-style prompting~\citep{yao2022react} and retrieves the top-3 skills from the bank via a hybrid scorer that combines BM25~\citep{robertson2009probabilistic} with dense cosine similarity over the title, principle, and when\_to\_apply fields. We run the propose-verify loop for a fixed budget of 10 rounds. At each round, the \emph{Edit Planner} produces $K$ candidate banks, and the Pareto-aware selector uses a utility tolerance $\epsilon=0.03$ and breaks ties by hypervolume contribution over diversity and coverage. To reduce the cost of repeated counterfactual leave-one-out rollouts required by both \emph{Skill Diagnoser} and the $J_{\text{util}}$ computation, we employ a content-addressed counterfactual replay cache that is shared across stages and rounds; see Appendix~\ref{sec:app_cfcache} for details. All experiments run on a single 8$\times$A800 node.

\subsection{Main Results}

\paragraph{Comparison with baselines.}
Table~\ref{tab:main} compares {\ours} against ten training-free baselines on ALFWorld and WebShop, all running on the same frozen Qwen2.5-7B-Instruct worker. {\ours} achieves the highest overall average on ALFWorld at 59.0\%, surpassing the strongest skill-based method Skill-Pro by 9.7\% and ReAct by 27.8\%. Notably, it also outperforms the append-only Voyager by 12.0\%. Relative to ReAct, {\ours} improves on all six task types, with the largest gains of 47.7\% on Heat and 38.9\% on Cool. On WebShop, {\ours} achieves the best results on both metrics, with a Score of 59.3\% and a Succ.\ of 38.4\%. The substantial gain on Succ.\ indicates that the curated skills translate into precise purchases matching every required attribute, rather than near matches that earn only partial reward.

\paragraph{Cross-worker generalization.}
Table~\ref{tab:cross_basemodel} extends the evaluation to different worker backbones, including three Qwen open-source models~\citep{yang2025qwen3} and two closed-source models, GPT-4o~\citep{hurst2024gpt} and Gemini-2.5-Pro~\citep{comanici2025gemini}. {\ours} improves over the ReAct baseline on every backbone, with ALFWorld Avg.\ gains ranging from 20.2\% on Qwen3-30B-A3B-Instruct-2507 to 41.7\% on GPT-4o. The improvements are particularly pronounced on closed-source workers: on GPT-4o, ALFWorld Avg.\ rises from 46.4 to 88.1 and WebShop Succ.\ from 27.2 to 65.0, a gain of 37.8\%. Even after curation, the Qwen3-4B-Instruct-2507 worker reaches 60.4\%, approaching the 61.8\% performance of Gemini-2.5-Pro without curation, suggesting that a well-curated skill bank can partially compensate for model size.

\subsection{Analysis}

\subsubsection{Curation Loop Dynamics}
Figure~\ref{fig:dynamics} traces the dynamics of skill bank over the curation loop across five worker backbones on ALFWorld. The bank size contracts from larger cold-start values to compact banks within 9--10 rounds and stabilizes thereafter, indicating that the curation process converges across all workers rather than growing unboundedly. The test success rate rises throughout the loop on every worker, with small round-to-round fluctuations but clear overall gains. Even at round 1, where the bank only uses the cold-start output of the \emph{Skill Distiller}, the test success rate is already substantially above the ReAct baseline of each worker, confirming that the cold-start design provides a strong starting point that the subsequent curation loop further improves upon. Representative skills from the final curated bank for each benchmark are shown in Appendix~\ref{sec:appb}.
\subsubsection{Ablation Study}


\begin{table}[!t]
\centering
\caption{Bank transfer across workers on ALFWorld (Avg.\ SR, \%). \textbf{Bold} denotes best results. }
\label{tab:bank_transfer}
\footnotesize
\setlength{\tabcolsep}{3pt}
\begin{tabular}{c ccc}
\toprule
& \multicolumn{3}{c}{\textbf{Agent Worker}} \\
\cmidrule(lr){2-4}
\textbf{Bank Source} &
  \makecell{Qwen3-4B\\Instruct-2507} &
  \makecell{Qwen3-30B-A3B\\Instruct-2507} &
  GPT-4o \\
\midrule
\textit{no bank (ReAct)}      & 38.8 & 55.2 & 46.4 \\
\midrule
\makecell{Qwen3-4B\\Instruct-2507}        & \cellcolor{gray!15}\textbf{60.4} & 71.6 & 78.4 \\
\makecell{Qwen3-30B-A3B\\Instruct-2507}   & 52.2 & \cellcolor{gray!15}\textbf{75.4} & 82.8 \\
GPT-4o                                     & 53.7 & 73.9 & \cellcolor{gray!15}\textbf{88.1} \\
\bottomrule
\end{tabular}
\end{table}

\paragraph{Bank transferability.}
Table~\ref{tab:bank_transfer} examines whether a curated bank for one worker remains effective when applied to a different worker at inference time. We curate three skill banks with {\ours}, each under a different source worker, and apply each bank to all three inference workers. Every cell substantially exceeds the ReAct baseline without retrieval. Diagonal cells, where each worker retrieves from its own bank, yield the best result for that worker at 60.4\%, 75.4\%, and 88.1\%, indicating that curation specializes the bank to its source worker. Cells away from the diagonal remain close behind in both directions: a bank curated by the smallest Qwen3-4B-Instruct-2507 worker lifts GPT-4o from 46.4\% to 78.4\%, only 9.7\% below GPT-4o's own result; conversely, a bank curated by GPT-4o lifts Qwen3-4B-Instruct-2507 worker from 38.8\% to 53.7\%, only 6.7\% below its own result. This suggests that the curated skills encode transferable procedural knowledge rather than artifacts of a particular worker, so a skill bank can be reused across workers in either direction with little loss.


\begin{table}[!t]
\centering
\caption{Ablation on the objective formulation. Each row uses a different subset of three objectives, while keeping the rest of the curation pipeline fixed. \textbf{Bold} denotes best results.}
\label{tab:ablation_objective}
\small
\setlength{\tabcolsep}{5pt}
\begin{tabular}{l ccc}
\toprule
& \textbf{ALFWorld} & \multicolumn{2}{c}{\textbf{WebShop}} \\
\cmidrule(lr){2-2} \cmidrule(lr){3-4}
\textbf{Objectives} & Avg. & Score & Succ. \\
\midrule
$J_{\text{util}}$                                          & 45.8 & 48.2 & 28.4 \\
$J_{\text{util}} + J_{\text{div}}$                         & 51.4 & 47.6 & 36.5 \\
$J_{\text{util}} + J_{\text{cov}}$                         & 52.6 & 51.0 & 37.6 \\
\midrule
$J_{\text{util}} + J_{\text{div}} + J_{\text{cov}}$  & \textbf{59.0} & \textbf{59.3} & \textbf{38.4} \\
\bottomrule
\end{tabular}
\end{table}

\paragraph{Effect of objective.}
Table~\ref{tab:ablation_objective} ablates the contribution of each objective by varying which terms enter the Pareto-aware selector while keeping curation pipeline fixed. Using $J_{\text{util}}$ alone yields a success rate of 45.8\% on ALFWorld and 28.4\% on WebShop, trailing the full objective by 13.2 and 10.0\% respectively. Compared to using $J_{\text{util}}$ alone, adding $J_{\text{div}}$ or adding $J_{\text{cov}}$ improves the success rate. The two objectives are complementary: $J_{\text{div}}$ promotes diversity and prevents redundancy in skill bank, while $J_{\text{cov}}$ penalizes banks that bloat with dormant or rarely retrieved skills. Combining both reaches the best results on both benchmarks.

\begin{table}[!t]
\centering
\caption{Ablation on the edit operations. Each row uses a different subset of the three edit operations. \textbf{Bold} denotes best results.}
\label{tab:ablation_edit}
\small
\setlength{\tabcolsep}{4pt}
\begin{tabular}{ccc c cc}
\toprule
\multicolumn{3}{c}{\textbf{Edit Operation}} & \textbf{ALFWorld} & \multicolumn{2}{c}{\textbf{WebShop}} \\
\cmidrule(lr){1-3} \cmidrule(lr){4-4} \cmidrule(lr){5-6}
\textsc{Add} & \textsc{Rewrite} & \textsc{Remove} & Avg. & Score & Succ. \\
\midrule
\checkmark &            &            & 47.0 & 50.0 & 26.4 \\
\checkmark & \checkmark &            & 53.5 & 51.0 & 37.6 \\
\checkmark &            & \checkmark & 48.3 & 49.5 & 34.5 \\
\midrule
\checkmark & \checkmark & \checkmark & \textbf{59.0} & \textbf{59.3} & \textbf{38.4} \\
\bottomrule
\end{tabular}
\end{table}

\paragraph{Effect of edit operations.}
Table~\ref{tab:ablation_edit} ablates the three edit operations available to the \emph{Edit Planner}. With \textsc{Add} alone, the bank grows monotonically without ever pruning or refining, leaving dormant or harmful skills to crowd out retrieval slots; the success rate reaches only 47.0\% on ALFWorld and 26.4\% on WebShop. The \textsc{Add}+\textsc{Rewrite} setting lifts ALFWorld to 53.5\% because the majority of skills the Diagnoser flags are salvageable, and refinement recovers their value rather than discarding it. \textsc{Add}+\textsc{Remove} setting lifts ALFWorld only to 48.3\%, lower than \textsc{Add}+\textsc{Rewrite} because the operation is too coarse: every flagged skill must be discarded directly, including partially correct ones whose strategy could have been salvaged and refined by a small edit. Combining all three reaches 59.0\% on ALFWorld and 38.4\% on WebShop, indicating that the two operations play complementary roles: \textsc{Rewrite} is the default rescue path that handles the bulk of flagged skills, while \textsc{Remove} is reserved for the minority whose underlying strategy cannot be repaired by any edit.

\section{Conclusion}

We present SkillBrew, a training-free framework that curates skill banks for LLM agents through bank-level optimization. Instead of judging each skill independently or continuously adding new skills, SkillBrew treats the entire skill bank as the optimization object and balances utility, diversity, and coverage under a Pareto-aware selection process. Experiments on ALFWorld and WebShop show that this bank-level curation leads to more diverse, transferable, and effective skill repositories, improving agent performance across both open-source and closed-source workers. These results suggest that future self-improving agents should optimize skill banks as global knowledge systems, rather than manage them as isolated skills.

\section*{Limitations}
One limitation of our framework is that the agent remains frozen during skill bank optimization, leaving joint co-adaptation between the agent and the evolving bank unexplored. Future work could investigate recursive co-training or reinforcement learning frameworks that jointly optimize both components. In addition, although our content-addressed cache substantially reduces the cost of counterfactual rollouts, the method still relies on repeated rollouts and leave-one-out replay, which may become expensive at larger scales; a more efficient method could further improve scalability. Finally, the current formulation assumes relatively static task distributions, and extending bank curation to open-world and continually shifting environments remains an important direction. Nevertheless, we believe this work introduces a new perspective: skill banks should be treated not as append-only memory repositories, but as globally optimized and continually refined objects, opening new directions for scalable and lifelong agent systems.

\bibliography{main}

\appendix

\section{Implementation Details}
\label{sec:appa}

\subsection{Datasets}

\paragraph{ALFWorld.}
ALFWorld~\citep{shridhar2020alfworld} is a text-based embodied benchmark that spans six household task families. We randomly sample 200 episodes from the official \texttt{train} split as the support set $\mathcal{D}_{\text{support}}$, and use the full \texttt{valid\_seen} split as the query set $\mathcal{D}_{\text{query}}$ and the full \texttt{valid\_unseen} split as the test set. Sampling uses random seed 42.

\paragraph{WebShop.}
WebShop~\citep{yao2022webshop} is a simulated online shopping environment with a fixed product catalog and one-million-scale natural language goals; the standard test split contains 500 goals, which we use as the test set. We additionally sample 500 and 250 goals from WebShop training and evaluation pool (goal ids $\geq 500$) as $\mathcal{D}_{\text{support}}$ and $\mathcal{D}_{\text{query}}$ respectively, with seed 42. The support, query, and test sets are pairwise disjoint.

\subsection{LLM Settings}

The agent worker is prompted in ReAct style~\citep{yao2022react}. On ALFWorld, the worker is given a budget of 50 steps per episode and conditions on the full history at every generation step. On WebShop, the worker is given 15 steps per episode and conditions on the most recent one step as history. The maximum generation length per step is 2{,}048 tokens for all agent workers. For \emph{Skill Distiller}, \emph{Skill Diagnoser}, and \emph{Edit Planner}, we all use GPT-5.4~\citep{openai2026gpt54}. Their prompts are shown in Appendix~\ref{sec:appc}.

\subsection{Skill Retrieval}

The agent retrieves the top-3 most relevant skills from skill bank using a hybrid scorer that combines BM25~\citep{robertson2009probabilistic} with dense cosine similarity. Concretely, the score of a skill $s$ relative to a query $q$ is
\[
\text{score}(q, s) = 0.30 \cdot \overline{\text{BM25}}(q, s) + 0.70 \cdot \cos(\mathbf{e}_q, \mathbf{e}_s),
\]
where $\overline{\text{BM25}}$ denotes min-max normalized BM25 scores. The skill text used for retrieval is the combination of three fields: \texttt{title}, \texttt{principle}, and \texttt{when\_to\_apply}. Skills with a combined score below 0.30 are dropped from the retrieval result. We use text-embedding-3-small for generating skill embeddings.

\subsection{Curation Loop and Pareto-Aware Selector}

We run the propose-then-verify loop for a fixed budget of 10 rounds. Within each round, the \emph{Skill Distiller}, \emph{Skill Diagnoser}, and \emph{Edit Planner} are invoked in this order, and each candidate bank proposed by the Edit Planner is evaluated by rolling out the worker on $\mathcal{D}_{\text{query}}$ to estimate $J_{\text{util}}$, $J_{\text{div}}$, and $J_{\text{cov}}$. The Pareto-aware selector uses a hypervolume contribution criterion with lex priority on $J_{\text{util}}$: candidates whose $J_{\text{util}}$ falls within a tolerance of $\epsilon = 0.03$ of the highest feasible $J_{\text{util}}$ are admitted to the tie pool, and ties are broken by hypervolume contribution over $J_{\text{div}}$ and $J_{\text{cov}}$. The previous bank is augmented to the candidate set as a null candidate. If it wins the lex-Pareto tie-break by hypervolume contribution, the bank carries forward unchanged to the next round.

\subsection{Counterfactual Replay Cache}
\label{sec:app_cfcache}

The propose-then-verify loop requires counterfactual leave-one-out rollouts in two places: on support set $\mathcal{D}_{\text{support}}$, where the Skill Diagnoser compares with and without each retrieved skill to issue \textsc{Keep} / \textsc{Rewrite} / \textsc{Remove} verdicts, and on query set $\mathcal{D}_{\text{query}}$, where each candidate bank's $J_{\text{util}}$ is computed by the same leave-one-out procedure. 

To improve efficiency across both stages, we maintain a single content-addressed cache shared by all rollouts. Each entry is keyed by a SHA-256 hash of the task id and the canonical content (\texttt{title}, \texttt{principle}, \texttt{when\_to\_apply}) of every retrieved skill, ordered by retrieval rank, so any change visible to the worker is also visible to the key. Entries also carry a version tag derived from the worker configuration; whenever the worker model or its prompt template changes, all cached entries are silently invalidated, preventing stale rollouts from polluting either the Diagnoser verdicts or the $J_{\text{util}}$ signal. 

Because the $K$ candidate banks proposed by the Edit Planner within a round differ by only a few edits, and the winning bank further carries forward to the next round, most leave-one-out retrievals coincide across candidates and across rounds. We observe a cache hit rate above 60\% from the second round onwards, reducing the cost of counterfactual rollouts.

\section{Skill Examples}
\label{sec:appb}
\begin{figure*}
    \centering
    \includegraphics[width=1\linewidth]{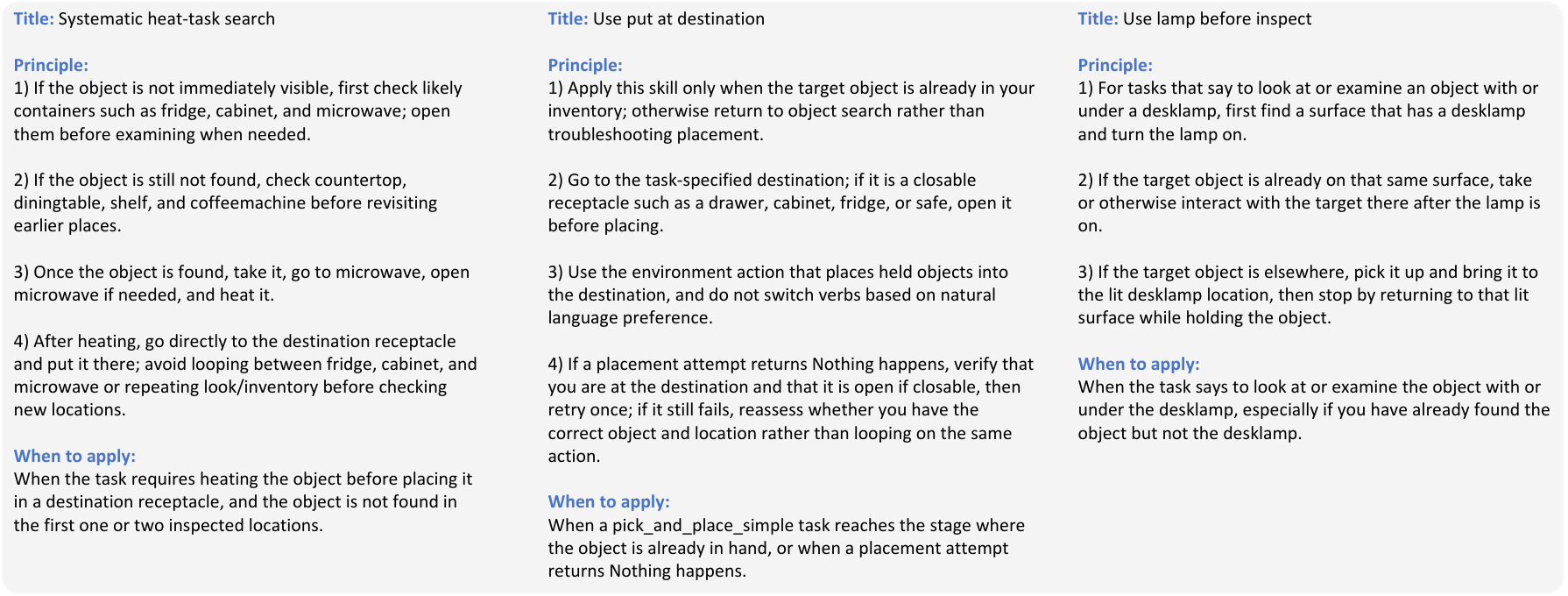}
    \caption{Example skills from the curated bank on ALFWorld.}
    \label{fig:casealfworld}
\end{figure*}

\begin{figure*}
    \centering
    \includegraphics[width=1\linewidth]{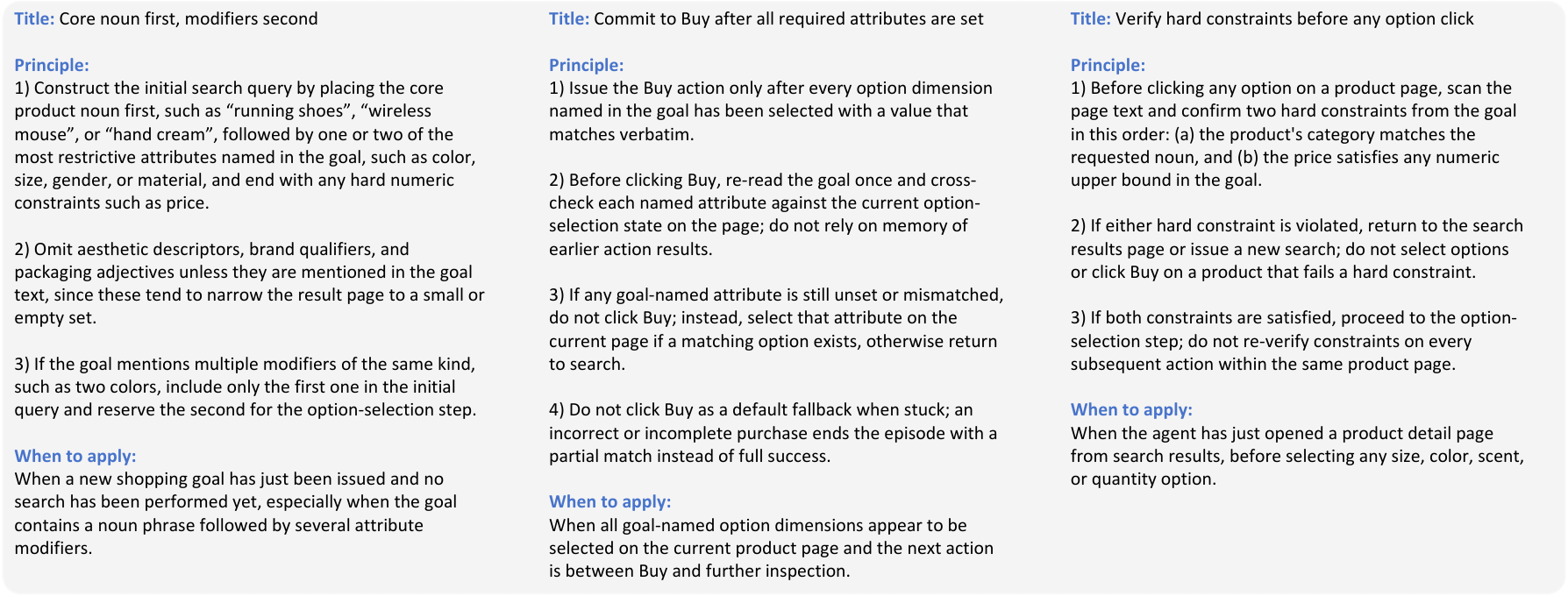}
    \caption{Example skills from the curated bank on WebShop.}
    \label{fig:casewebshop}
\end{figure*}

\begin{figure*}
    \centering
    \includegraphics[width=1\linewidth]{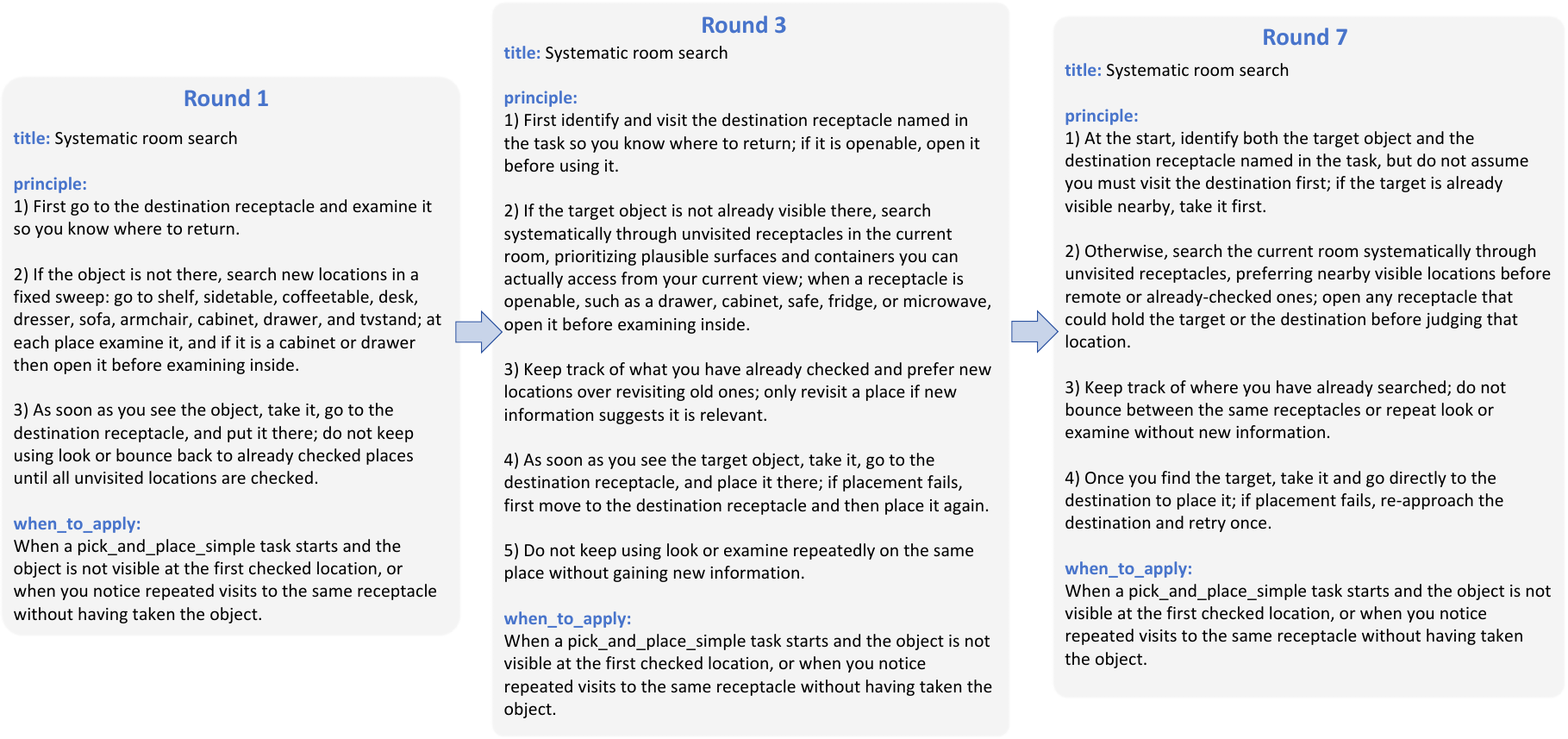}
    \caption{An example of \textsc{Rewrite}.}
    \label{fig:rewrite_case}
\end{figure*}

Figure~\ref{fig:casealfworld} and Figure~\ref{fig:casewebshop} present representative skills from the final curated bank on ALFWorld and WebShop, respectively; Figure~\ref{fig:rewrite_case} additionally traces a single skill before and after a \textsc{Rewrite} verdict from the \emph{Skill Diagnoser}.

\section{Prompts}
\label{sec:appc}
\begin{figure*}
    \centering
    \includegraphics[width=1\linewidth]{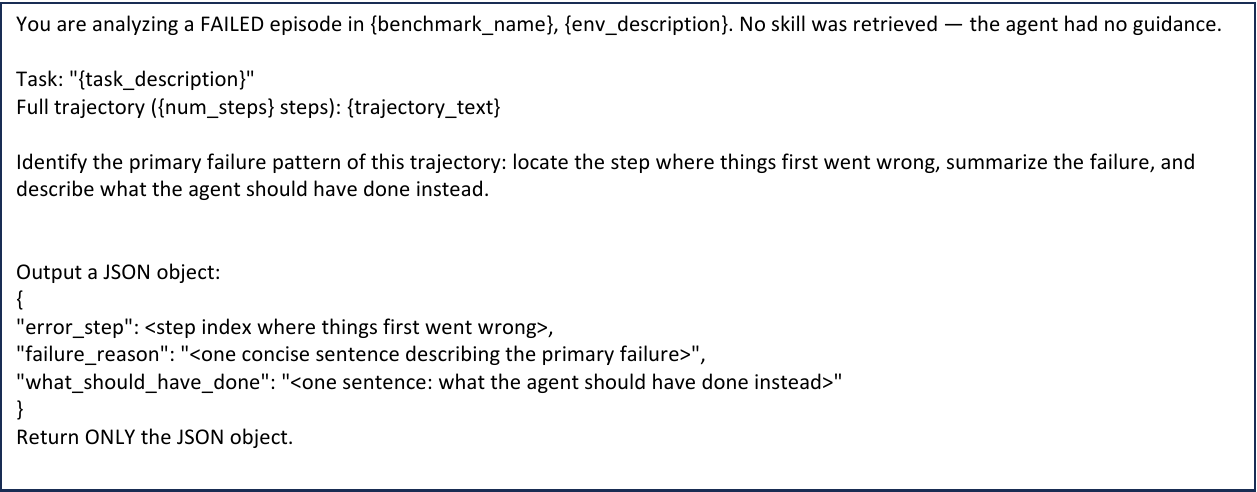}
    \caption{Prompt of the \emph{Skill Distiller} for failure analysis.}
    \label{fig:prompt_distiller_failure}
\end{figure*}

\begin{figure*}
    \centering
    \includegraphics[width=1\linewidth]{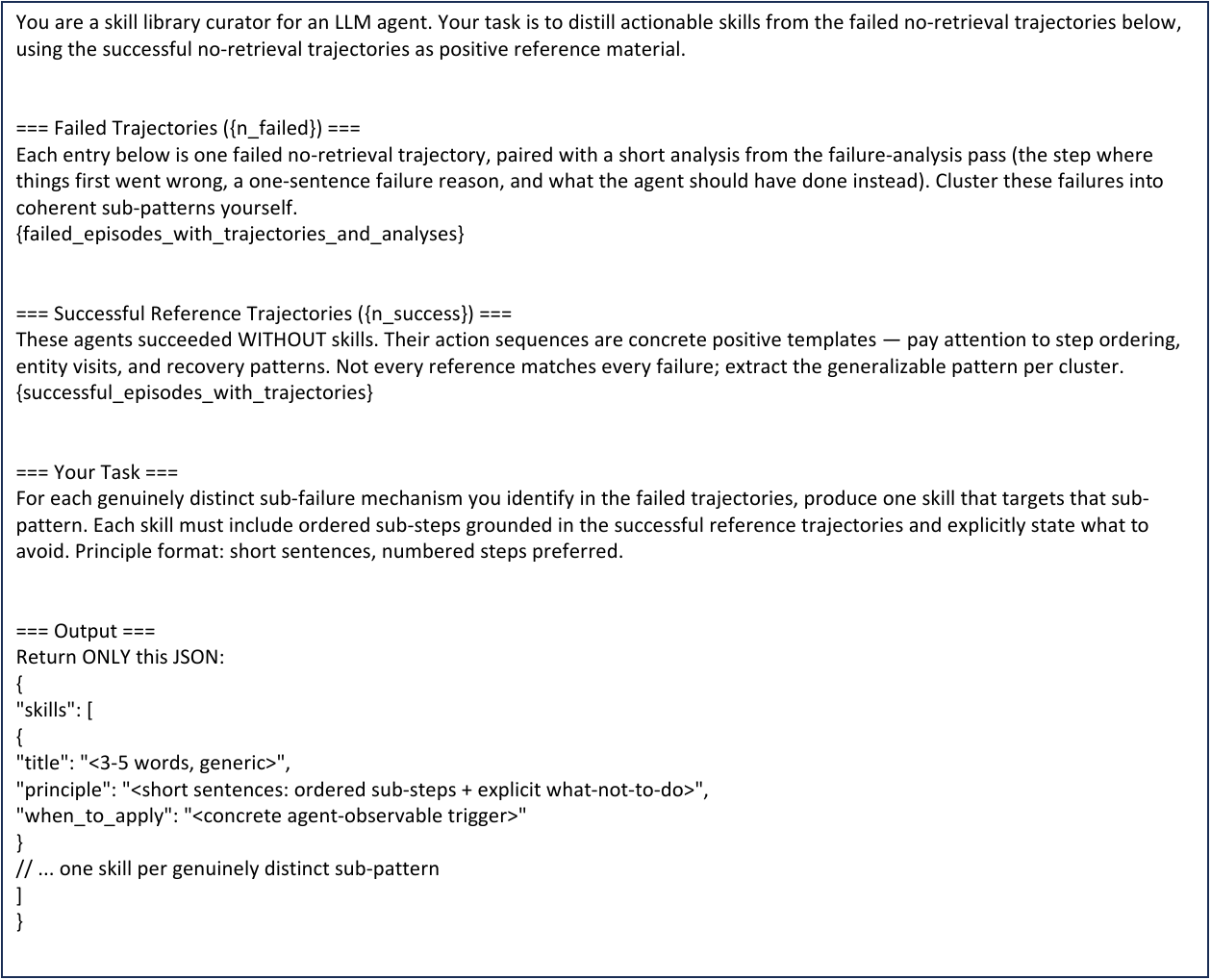}
    \caption{Prompt of the \emph{Skill Distiller} for producing candidate skills.}
    \label{fig:prompt_distiller_synthesis}
\end{figure*}

\begin{figure*}
    \centering
    \includegraphics[width=1\linewidth]{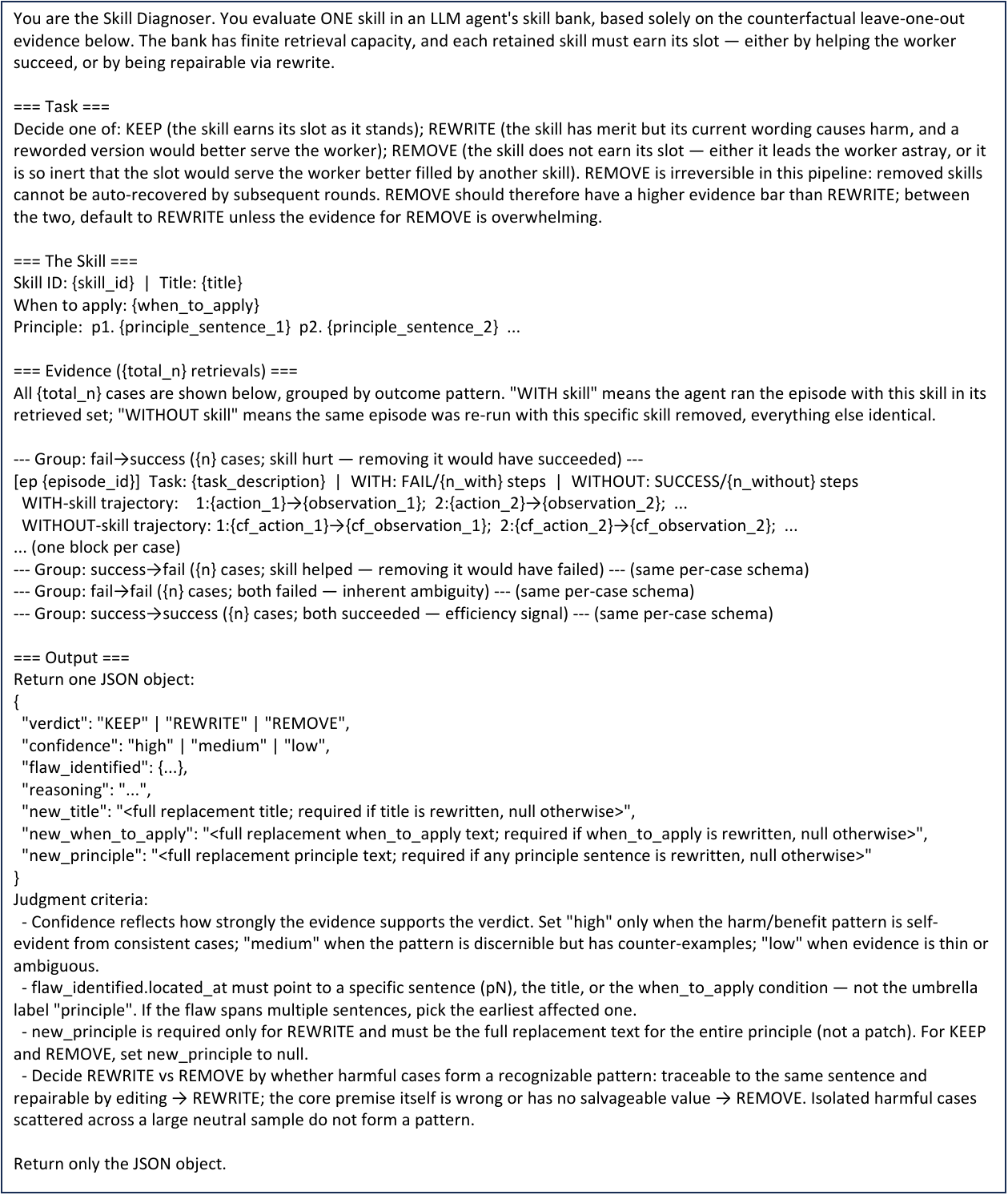}
    \caption{Prompt of the \emph{Skill Diagnoser}}
    \label{fig:prompt_diagnoser}
\end{figure*}

\begin{figure*}
    \centering
    \includegraphics[width=1\linewidth]{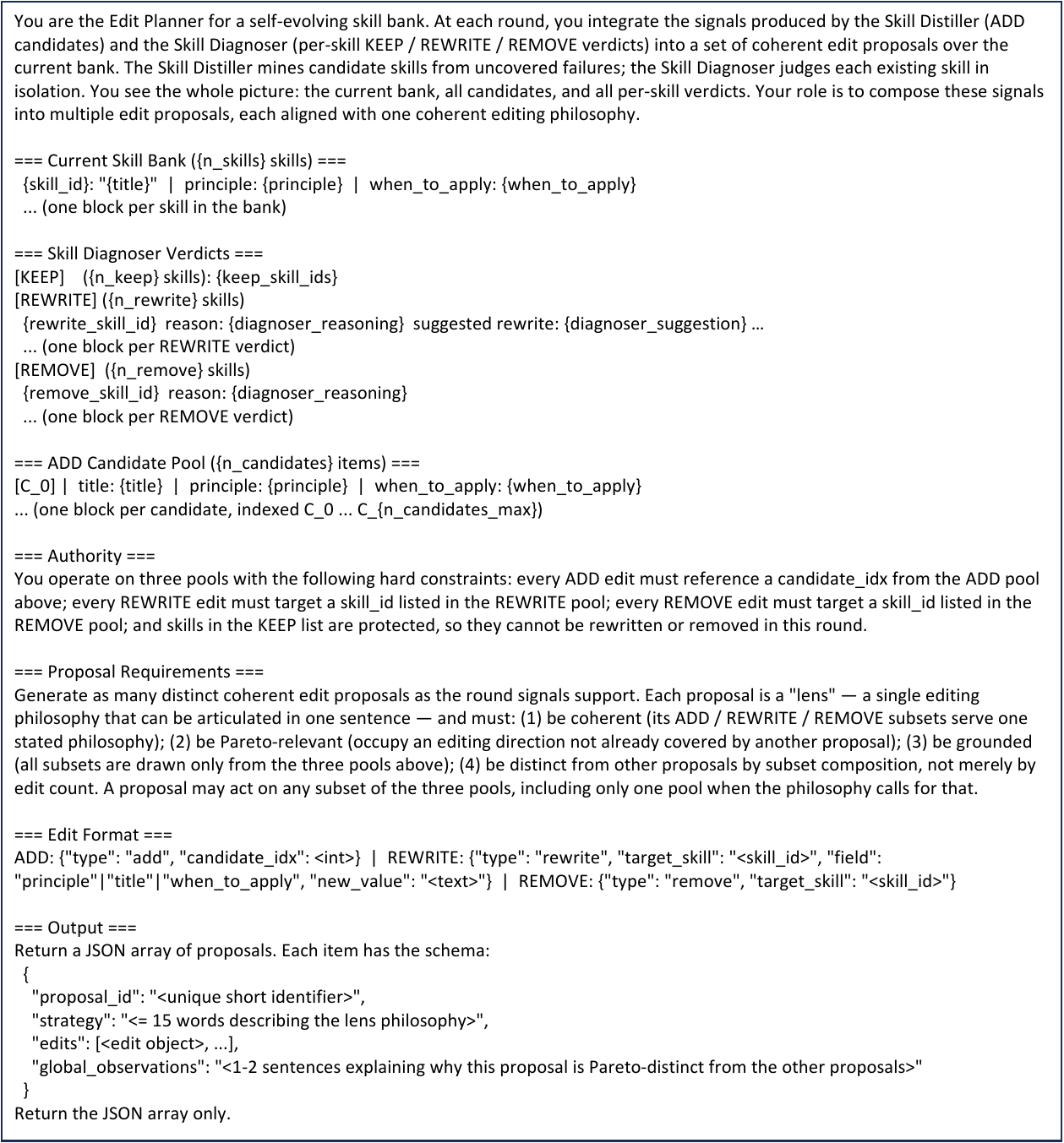}
    \caption{Prompt of the \emph{Edit Planner}}
    \label{fig:prompt_planner}
\end{figure*}

Figures~\ref{fig:prompt_distiller_failure} and~\ref{fig:prompt_distiller_synthesis} show the two-stage prompt used by the \emph{Skill Distiller}: the first analyzes failure trajectories to identify missing capabilities, while the second synthesizes candidate \textsc{Add} skills targeting those gaps. Figure~\ref{fig:prompt_diagnoser} shows the prompt of the \emph{Skill Diagnoser}, which assigns \textsc{Keep}, \textsc{Rewrite}, or \textsc{Remove} verdicts to each retrieved skill from its factual and counterfactual outcomes. Figure~\ref{fig:prompt_planner} shows the prompt of the \emph{Edit Planner}. Rather than mechanically applying each edit verdict, the planner conditions on the full current bank together with all candidate edit pools, reasoning at the bank level about which combinations of edits compose into $K$ candidate banks.

\clearpage

\begin{algorithm}[t]
\small
\caption{Inner Loop: Skill Edit Proposal from Support Trajectories}
\label{alg:inner_loop}
\SetAlgoNoLine
\DontPrintSemicolon
\SetAlgoNoEnd
\SetKwInput{KwIn}{\textbf{Input}}
\SetKwInput{KwOut}{\textbf{Output}}

\SetKwFunction{Rollout}{Rollout}
\SetKwFunction{Retrieve}{Retrieve}
\SetKwFunction{Distill}{Distill}
\SetKwFunction{Diagnose}{Diagnose}
\SetKwFunction{Replay}{Replay}
\SetKwFunction{PlanEdit}{PlanEdit}

\KwIn{Current skill bank $\mathcal{B}^{(i)}$, support set $\mathcal{D}_{\text{support}}$, frozen agent $\pi_{\text{agent}}$, retriever $\textsc{ret}$, number of candidates $K$}
\KwOut{Candidate bank set $\mathcal{K}^{(i)} = \{\mathcal{B}^{(i+1)}_k\}_{k=1}^{K}$}

\algcmt{Roll out and partition trajectories by outcome and retrieval status}

$\mathcal{T}^{+}_{\bar{R}} \leftarrow \emptyset$;
$\mathcal{T}^{-}_{\bar{R}} \leftarrow \emptyset$;
$\mathcal{T}^{+}_{R} \leftarrow \emptyset$;
$\mathcal{T}^{-}_{R} \leftarrow \emptyset$\;

\ForEach{$t \in \mathcal{D}_{\text{support}}$}{

    $R_t \leftarrow \textsc{ret}(\mathcal{B}^{(i)}, t)$;\quad
    $\tau_t \leftarrow \Rollout(\pi_{\text{agent}}, t, R_t)$\;

    \uIf{$\textsc{Succ}(\tau_t)$ \textbf{and} $R_t = \emptyset$}{
        $\mathcal{T}^{+}_{\bar{R}} \leftarrow \mathcal{T}^{+}_{\bar{R}} \cup \{\tau_t\}$\;
    }
    \uElseIf{$\textsc{Fail}(\tau_t)$ \textbf{and} $R_t = \emptyset$}{
        $\mathcal{T}^{-}_{\bar{R}} \leftarrow \mathcal{T}^{-}_{\bar{R}} \cup \{\tau_t\}$\;
    }
    \uElseIf{$\textsc{Succ}(\tau_t)$ \textbf{and} $R_t \neq \emptyset$}{
        $\mathcal{T}^{+}_{R} \leftarrow \mathcal{T}^{+}_{R} \cup \{(t, R_t, \tau_t)\}$\;
    }
    \Else{
        $\mathcal{T}^{-}_{R} \leftarrow \mathcal{T}^{-}_{R} \cup \{(t, R_t, \tau_t)\}$\;
    }
}

\algcmt{Distill \textsc{Add} candidates from no-retrieval failures, with no-retrieval successes as positive references}

$\mathcal{E}_{\text{add}} \leftarrow \Distill(\mathcal{T}^{-}_{\bar{R}}, \mathcal{T}^{+}_{\bar{R}})$\;

\algcmt{Collect leave-one-out evidence pairs per retrieved skill}

$\mathcal{P}_{s} \leftarrow \emptyset$ \textbf{for each} $s$\;

\ForEach{$(t, R_t, \tau_t) \in \mathcal{T}^{+}_{R} \cup \mathcal{T}^{-}_{R}$}{
    \ForEach{$s \in R_t$}{
        $\tilde{\tau}_{t \setminus s} \leftarrow \Replay(\pi_{\text{agent}}, t, R_t \setminus \{s\})$\;
        $\mathcal{P}_{s} \leftarrow \mathcal{P}_{s} \cup \{(\tau_t, \tilde{\tau}_{t \setminus s})\}$\;
    }
}

\algcmt{Diagnose each skill once over its full pool of counterfactual pairs}

$\mathcal{E}_{\text{keep}} \leftarrow \emptyset$;
$\mathcal{E}_{\text{rewrite}} \leftarrow \emptyset$;
$\mathcal{E}_{\text{remove}} \leftarrow \emptyset$\;

\ForEach{$s$ \textbf{with} $\mathcal{P}_{s} \neq \emptyset$}{

    $v_s \leftarrow \Diagnose(s, \mathcal{P}_{s})$\;

    \uIf{$v_s = \textsc{Keep}$}{
        $\mathcal{E}_{\text{keep}} \leftarrow \mathcal{E}_{\text{keep}} \cup \{s\}$\;
    }
    \uElseIf{$v_s = \textsc{Rewrite}$}{
        $\mathcal{E}_{\text{rewrite}} \leftarrow \mathcal{E}_{\text{rewrite}} \cup \{(s, \mathcal{P}_{s})\}$\;
    }
    \Else{
        $\mathcal{E}_{\text{remove}} \leftarrow \mathcal{E}_{\text{remove}} \cup \{s\}$\;
    }
}

\algcmt{Compose edits into candidate banks; \textsc{Keep} set is passed as protected slots}

$\mathcal{K}^{(i)} \leftarrow \PlanEdit
\left(
\mathcal{B}^{(i)},
\mathcal{E}_{\text{add}},
\mathcal{E}_{\text{rewrite}},
\mathcal{E}_{\text{remove}},
\mathcal{E}_{\text{keep}},
K
\right)$\;

\Return{$\mathcal{K}^{(i)}$}\;

\end{algorithm}

\begin{algorithm}[t]
\small
\caption{Outer Loop: Pareto-Aware Skill Bank Curation}
\label{alg:outer_loop}
\SetAlgoNoLine
\DontPrintSemicolon
\SetAlgoNoEnd
\SetKwInput{KwIn}{\textbf{Input}}
\SetKwInput{KwOut}{\textbf{Output}}

\SetKwFunction{ColdStart}{ColdStart}
\SetKwFunction{InnerLoop}{InnerLoop}
\SetKwFunction{EvalUtil}{Eval$_{\text{util}}$}
\SetKwFunction{EvalDiv}{Eval$_{\text{div}}$}
\SetKwFunction{EvalCov}{Eval$_{\text{cov}}$}
\SetKwFunction{ParetoFront}{ParetoFront}
\SetKwFunction{Hypervolume}{Hypervolume}

\KwIn{Support set $\mathcal{D}_{\text{support}}$, query set $\mathcal{D}_{\text{query}}$, frozen agent $\pi_{\text{agent}}$, retriever $\textsc{ret}$, max rounds $T$, number of candidates $K$, utility tolerance $\epsilon$}
\KwOut{Curated skill bank $\mathcal{B}^{\star}$}

\algcmt{Initialize skill bank from no-retrieval trajectories via the Skill Distiller}

$\mathcal{B}^{(0)} \leftarrow \ColdStart(\pi_{\text{agent}}, \mathcal{D}_{\text{support}})$\;

\For{$i=0$ \KwTo $T-1$}{

    \algcmt{Propose candidate banks from support-set evidence}

    $\mathcal{K}^{(i)} \leftarrow
    \InnerLoop
    \left(
    \mathcal{B}^{(i)},
    \mathcal{D}_{\text{support}},
    \pi_{\text{agent}},
    \textsc{ret},
    K
    \right)$\;

    \algcmt{Include the current bank as a null candidate to enable a no-edit decision}

    $\mathcal{C}^{(i)} \leftarrow \mathcal{K}^{(i)} \cup \{\mathcal{B}^{(i)}\}$\;

    \algcmt{Evaluate utility, diversity, and coverage on query set}

    \ForEach{$\mathcal{B} \in \mathcal{C}^{(i)}$}{
        $J_{\text{util}}(\mathcal{B}) \leftarrow \EvalUtil(\mathcal{B}, \mathcal{D}_{\text{query}}, \pi_{\text{agent}}, \textsc{ret})$ \tcp*{requires LOO rollouts}
        $J_{\text{div}}(\mathcal{B}) \leftarrow \EvalDiv(\mathcal{B})$ \tcp*{structural, no rollouts}
        $J_{\text{cov}}(\mathcal{B}) \leftarrow \EvalCov(\mathcal{B}, \mathcal{D}_{\text{query}}, \textsc{ret})$ \tcp*{retrieval only}
        $\Phi(\mathcal{B}) \leftarrow \big(J_{\text{util}}(\mathcal{B}), J_{\text{div}}(\mathcal{B}), J_{\text{cov}}(\mathcal{B})\big)$\;
    }

    \algcmt{Construct the non-dominated Pareto front}

    $\mathcal{F}^{(i)} \leftarrow
    \ParetoFront
    \left(
    \{(\mathcal{B}, \Phi(\mathcal{B})) : \mathcal{B} \in \mathcal{C}^{(i)}\}
    \right)$\;

    \algcmt{Prioritize utility, then break ties by diversity--coverage hypervolume}

    $u_{\max} \leftarrow
    \max_{\mathcal{B} \in \mathcal{F}^{(i)}} J_{\text{util}}(\mathcal{B})$\;

    $\mathcal{M}^{(i)} \leftarrow
    \left\{
    \mathcal{B} \in \mathcal{F}^{(i)}
    :
    J_{\text{util}}(\mathcal{B}) \geq u_{\max} - \epsilon
    \right\}$\;

    $\mathcal{B}^{(i+1)} \leftarrow
    \arg\max_{\mathcal{B} \in \mathcal{M}^{(i)}}
    \Hypervolume
    \left(
    J_{\text{div}}(\mathcal{B}),
    J_{\text{cov}}(\mathcal{B})
    \right)$
    \tcp*{if the null candidate wins, $\mathcal{B}^{(i+1)} = \mathcal{B}^{(i)}$}\;
}
$\mathcal{B}^{\star} \leftarrow \mathcal{B}^{(T)}$\;
\Return{$\mathcal{B}^{\star}$}\;

\end{algorithm}

\end{document}